\pdfoutput=1

\documentclass[11pt,table]{article}

\usepackage[]{EMNLP2022}

\usepackage{xcolor}
\usepackage{times}
\usepackage{latexsym}
\usepackage{graphicx}
\usepackage{booktabs} 
\usepackage{siunitx}
\usepackage{enumitem}
\setlist{nosep}
\usepackage{multirow}

\usepackage[T1]{fontenc}

\usepackage[utf8]{inputenc}

\usepackage{microtype}

\usepackage{inconsolata}

\usepackage{siunitx}
\newcommand{\crossmodal}{\textsc{\small XM3600}}
\newcommand{\xmmt}[1]{\textsc{\small XM3600}-#1-MT}
\newcommand{\xmsix}{\textsc{\small XM600}}
\newcommand{\cocodev}{\textsc{\small COCO-dev}}
\newcommand{\langspop}{Arabic~(ar), Chinese-Simplified~(zh), Croatian~(hr), Czech~(cs), Danish~(da), Dutch~(nl), Filipino~(fil), Finnish~(fi), French~(fr), German~(de), Greek~(el), Hebrew~(he), Hindi~(hi), Hungarian~(hu), Indonesian~(id), Italian~(it), Japanese~(ja), Korean~(ko), Norwegian~(no), Persian~(fa), Polish~(pl), Portuguese~(pt), Romanian~(ro), Russian~(ru), Spanish~(es), Swedish~(sv), Thai~(th), Turkish~(tr), Ukrainian~(uk), Vietnamese~(vi)}
\newcommand{\lthirty}{$\mathcal L30$}

\newcommand{\langslowres}{Bengali~(bn), Cusco Quechua~(quz), Maori~(mi), Swahili~(sw), Telugu~(te)}
\newcommand{\lfive}{$\mathcal L5$}
\newcommand{\lfour}{{$\mathcal L$}\textsc{\small core}}
\newcommand{\ofour}{{$\mathcal O$}\textsc{\small core}}
\newcommand{\langsfour}{Chinese-Simplified~(zh), English~(en), Hindi~(hi), Spanish~(es)}

\newcommand{\lall}{{$\mathcal L$}\textsc{\small all}}
\newcommand{\oall}{{$\mathcal O$}\textsc{\small all}}
\newcommand{\leval}{{$\mathcal L$}\textsc{\small ext}}
\newcommand{\oeval}{{$\mathcal O$}\textsc{\small ext}}
\newcommand{\langseval}{Arabic~(ar), Bengali~(bn), Croatian~(hr), Czech~(cs), Danish~(da), Dutch~(nl), Filipino~(fil), Finnish~(fi), French~(fr), German~(de), Greek~(el), Hebrew~(he), Hungarian~(hu), Indonesian~(id), Italian~(it), Japanese~(ja), Korean~(ko), Norwegian~(no), Persian~(fa), Polish~(pl), Portuguese~(pt), Romanian~(ro), Swahili~(sw), Swedish~(sv), Telugu~(te), Thai~(th), Turkish~(tr), Vietnamese~(vi)}

\newcommand{\sxsgain}{$\Delta$S$\times$S}
\newcommand{\wins}{\textsc{\small Wins}}
\newcommand{\losses}{\textsc{\small Losses}}
\newcommand{\numcaptions}{\num{261375}}

\newcommand{\baseBcccoco}{BB+CC}
\newcommand{\baseBcoco}{BB}
\newcommand{\basegcoco}{Bg}
\newcommand{\largegcoco}{Lg}

\title{Crossmodal-3600: A Massively Multilingual\\Multimodal Evaluation Dataset}

\author{
  Ashish V. Thapliyal,\ \ Jordi Pont-Tuset,\ \ Xi Chen,\ \ Radu Soricut\\
  Google Research \\
  \texttt{\normalsize\{asht,jponttuset,chillxichen,rsoricut\}@google.com}}

\begin{document}
\maketitle
\begin{abstract}
Research in massively multilingual image captioning has been severely hampered by a lack of
high-quality evaluation datasets.
In this paper we present the Crossmodal-3600 dataset (\crossmodal{} in short),
a geographically-diverse set of \num{3600} images annotated with human-generated
reference captions in \num{36} languages.
The images were selected from across the world, covering regions where the \num{36} languages are spoken,
and annotated with captions that achieve consistency in terms of style across all languages, while
avoiding annotation artifacts due to direct translation.
We apply this benchmark to model selection for massively multilingual image captioning models, and show
strong correlation results with human evaluations when using \crossmodal{} as golden
references for automatic metrics. 
\end{abstract}

\section{Introduction}

\begin{figure*}[ht!]
\vspace{-4mm}
\setlength{\fboxsep}{0pt}%
  \centering
  \begin{minipage}{0.45\linewidth}
  \centering
  \fbox{\includegraphics[width=0.9\linewidth]{figures/06c5f87181c30e00.jpeg}}\\[-1mm]
  {\tiny\it Source: \href{https://www.flickr.com/photos/briansolis/5129089526}{Porsche Museum, Stuttgart} by \href{https://www.flickr.com/people/briansolis/}{Brian Solis}.}
  \end{minipage}
  \begin{minipage}{0.54\linewidth}
  \includegraphics[width=\linewidth]{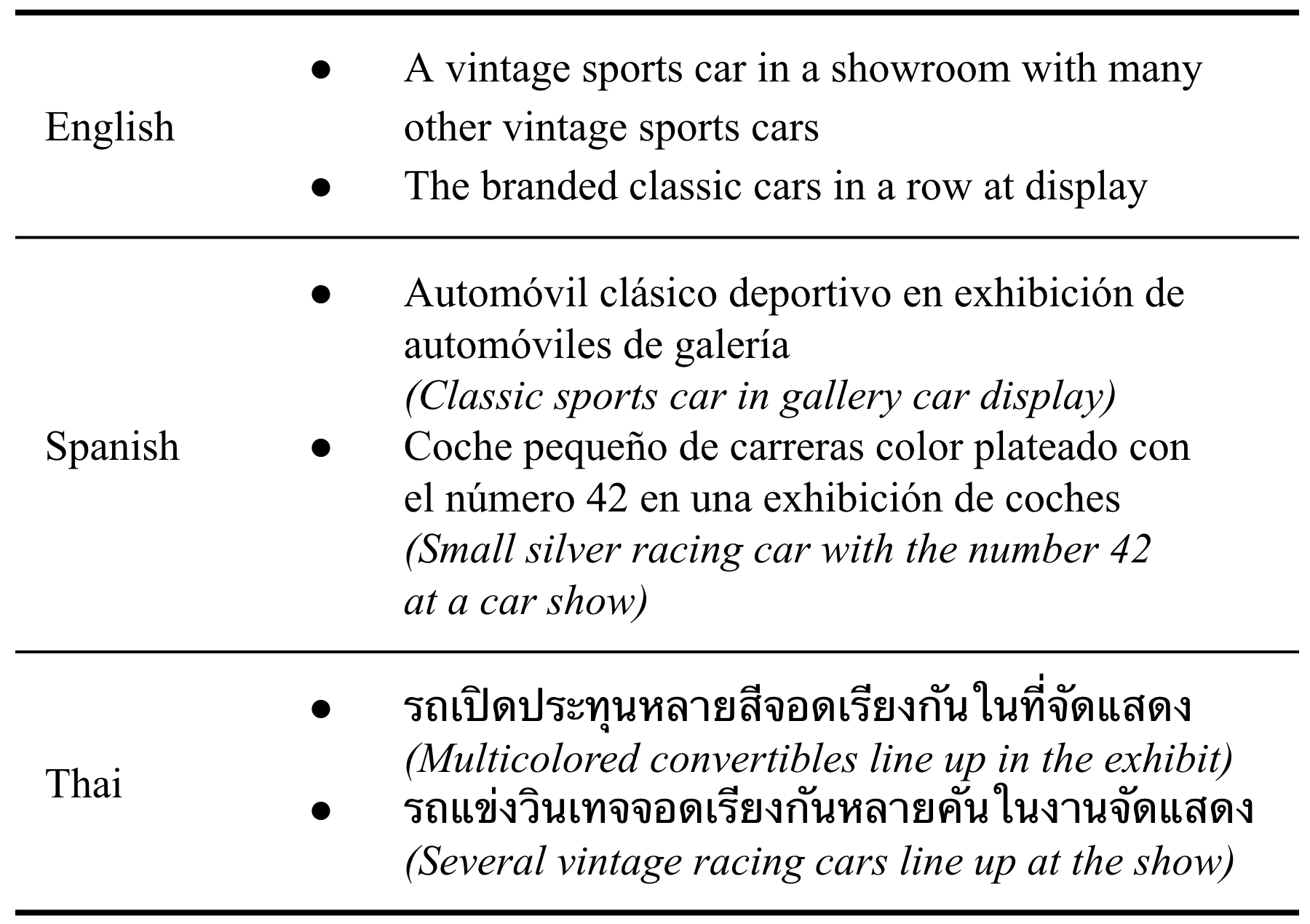}
  \end{minipage}
   \caption{\textbf{Sample captions in three different languages} (out of 36 -- see full list of captions in Appendix~\ref{app:additional_captions}), showcasing the creation of annotations that are consistent in style across languages, while being free of direct-translation artefacts (e.g.\ the Spanish ``number 42'' or the Thai ``convertibles'' would not be possible when directly translating from the English versions). }
  \label{fig:sample_captions}
\end{figure*}

Image captioning is the task of automatically generating a fluent natural language description for a given image.
This task is important for enabling accessibility for visually impaired users, and is a core task in multimodal research encompassing both vision and language modeling.
However, datasets for this task are primarily available in English \cite{young14tacl,chen2015microsoft,krishna2017visual,sharma2018conceptual,pont2020connecting}.
Beyond English, there are a few datasets such as Multi30K with captions in German~\cite{elliott:16}, French~\cite{elliott:17} and Czech~\cite{barrault:18}%
, but they are limited to only a few languages that cover a small fraction of the world's population, while featuring images that severely under-represent the richness and diversity of cultures from across the globe. 
These aspects have hindered research on image captioning for a wide variety of languages, and directly hamper deploying accessibility solutions for a large potential audience around the world.

Creating large training and evaluation datasets in multiple languages is a resource-intensive endeavor.
Recent works \cite{thapliyal-soricut-2020-cross} have shown that it is feasible to build multilingual image captioning models trained on machine-translated data (with English captions as the starting point).
This work also shows that the effectiveness of some of the most reliable automatic metrics for image captioning, such as CIDEr\footnote{"measures the similarity of a generated sentence against a set of ground truth sentences written by humans." quoted from ~\cite{cider}.}~\cite{cider} 
is severely diminished when applied to translated evaluation sets, resulting in poorer agreement with human evaluations compared to the English case.
As such, the current situation is that trustworthy model evaluation can only be based on extensive and expensive human evaluations.
However, such evaluations cannot usually be replicated across different research efforts, and therefore do not offer a fast and robust mechanism for model hill-climbing and comparison of multiple lines of research.

The proposed \crossmodal{} image captioning evaluation dataset provides a robust benchmark for multilingual image captioning, and can be reliably used to compare research contributions in this emerging field.
Our contributions are as follows:
(i) for human caption annotations, we have devised a protocol that allows  annotators for a specific target language to produce image captions in a style that is consistent across languages;
this protocol results in image-caption annotations that are free of direct translation artefacts, an issue that has plagued Machine Translation research for many years and is now well understood~\cite{freitag2020BLEUMB};
(ii) for image selection, we have devised an algorithmic approach to sample a set of 3600 geographically-diverse images from the Open Images Dataset~\cite{kuznetsova:2020}, aimed at creating a representative set of images from across the world;
(iii) for the resulting \crossmodal{} benchmark, we empirically measure its ability to rank image captioning model variations, and show that it provides high levels of agreement with human judgements, therefore validating its usefulness as a benchmark and alleviating the need for human judgement in the future.

\begin{figure*}[ht!]
\vspace{-4mm}
\setlength{\fboxsep}{0pt}%
\centering
\hspace{-2mm}
\begin{minipage}[t]{0.34\linewidth}
\centering
{\small English}
\fbox{\includegraphics[height=3.6cm]{figures/5ff986d798b66134.jpeg}}\\
\tiny\it Source: \href{https://www.flickr.com/photos/lodekka/5072748008}{APN59H 101010 CPS} by \href{https://www.flickr.com/people/lodekka/}{Chris Sampson}
\end{minipage}
\begin{minipage}[t]{0.34\linewidth}
\centering
{\small Swahili\vphantom{p}}
\fbox{\includegraphics[height=3.6cm]{figures/a91551a6e62da276.jpeg}}\\
\tiny\it Source: \href{https://www.flickr.com/photos/henrikpalm/7555980588}{Lens louse / Linslus} by \href{https://www.flickr.com/people/henrikpalm/}{Henrik Palm}
\end{minipage}
\hspace{-2pt}
\begin{minipage}[t]{0.3\linewidth}
\centering
{\small Telugu}
\fbox{\includegraphics[height=3.6cm]{figures/30bb25b4e1117219.jpeg}}\\
\tiny\it Source: \href{https://www.flickr.com/photos/14675798@N06/6901350577}{Garudan thookkam 02} by \href{https://www.flickr.com/people/14675798@N06/}{rojypala}
\end{minipage}\\[3mm]
\begin{minipage}[t]{0.34\linewidth}
\centering
{\small Cusco Quechua}
\fbox{\includegraphics[height=3.6cm]{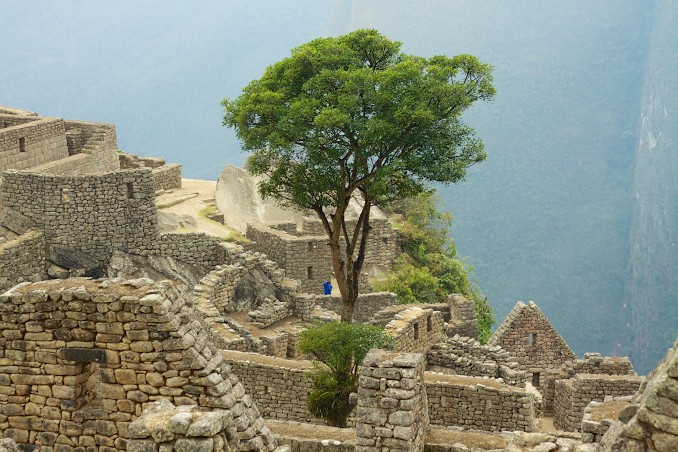}}
\tiny\it Source: \href{https://www.flickr.com/photos/mckaysavage/8296819497}{Peru - Machu Picchu 139} by \href{https://www.flickr.com/people/mckaysavage/}{McKay Savage}
\end{minipage}
\begin{minipage}[t]{0.34\linewidth}
\centering
{\small Filipino}
\fbox{\includegraphics[height=3.6cm]{figures/2911453cc61834d8.jpeg}}
\tiny\it Source: \href{https://www.flickr.com/photos/schoeters/2683433042}{Taal Lake Yacht Club} by \href{https://www.flickr.com/people/schoeters/}{Simon Schoeters}
\end{minipage}
\begin{minipage}[t]{0.3\linewidth}
\centering
{\small Chinese\vphantom{p}}
\fbox{\includegraphics[height=3.6cm]{figures/1d7c7320a508643f.jpeg}}
\tiny\it Source: \href{https://www.flickr.com/photos/rapidtravelchai/8639346056}{Shanghai Wangjia [...]} by \href{https://www.flickr.com/people/rapidtravelchai/}{Stefan Krasowski}
\end{minipage}
\vspace{-2mm}
\caption{A sample of images in the \crossmodal{} dataset, together with the language for which they have been selected. Overall, the images span regions over 36 different languages and 6 different continents.}
\label{fig:sample_images}
\end{figure*}

Fig.~\ref{fig:sample_captions} shows a few sample captions for an image in \crossmodal{} that exemplify point (i) above, and Fig.~\ref{fig:sample_images} shows the variety of cultural aspects captured by the image sampling approach from point (ii).
We provide detailed explanations and results for each of the points above in the rest of the paper.
We have released \crossmodal{} under a \href{https://creativecommons.org/licenses/by/4.0/}{CC-BY4.0} license at \href{https://google.github.io/crossmodal-3600/}{https://google.github.io/crossmodal-3600/}.

\section{The \crossmodal{} Dataset}
In this section, we describe the heuristics used for language and image selection, the design of the caption annotation process, caption statistics including quality, and annotator details.

\subsection{Language Selection}
In this section, we describe the heuristic used for selecting the languages. 
As a first step, we take a quantitative stance and choose 30 languages (\lthirty{}) roughly based on their percent of web content\footnote{\scriptsize\langspop.}.
As a second step, we consider an additional five languages (\lfive{}) \footnote{\scriptsize\langslowres.} to cover low-resource languages with many native speakers, or major native languages from continents that would not be covered otherwise. The protocol for caption annotation (Sec.~\ref{sec:caption-annotation}) has been applied to the resulting union of languages plus English, for a total of 36 languages.

\subsection{Image Selection}
\label{sec:image-selection}
In this section, we consider the heuristics used for selecting a geographically diverse set of images.
For each of the 36 languages, we select \num{100} images that, as far as it is possible for us to identify, are taken in an area where the given language is spoken.
The images are selected among those in the Open Images Dataset~\cite{kuznetsova:2020} that have GPS coordinates stored in their EXIF metadata.

Since there are many regions where more than one language is spoken, and given that some areas are not well covered by Open Images, we design an algorithm that maximizes the percentage of selected images taken in an area in which the assigned language is spoken. 
This is a greedy algorithm that starts the selection of images by the languages for which we have the smallest pool (e.g.\ Persian) and processes them in increasing order of their candidate image pool size.
Whenever there are not enough images in the area where a language is spoken, we have several back-off levels: (i) selecting from a country where the language is spoken; (ii) a continent where the language is spoken, and, as last resort, (iii) from anywhere in the world.

This strategy succeeds in providing our target number of 100 images from an appropriate region for most of the 36 languages except for Persian (where 14 continent-level images are used) and Hindi  (where all 100 images are at the global level because the in-region images are assigned to Bengali and Telugu).
We keep the region each image is selected from as part of our data annotation, so that future evaluations can choose to either evaluate on images relevant to particular regions of interest or on the entire dataset.

\subsection{Caption Annotation}
\label{sec:caption-annotation}
In this section we detail the design of the caption annotation process. 
For a massively multilingual benchmark such as \crossmodal{}, consistency in the style of the description language is critical, since language can serve multiple communication goals.
For a more in-depth discussion on these issues as they relate to image captions, we refer the reader to \cite{alikhani-etal-2020-cross}.
We borrow from their terminology, as it identifies coherence relations between image and captions such as \textsc{Visible}, \textsc{Meta}, \textsc{Subjective}, and \textsc{Story}. 
The goal for our caption annotation is to generate \textsc{Visible} image captions, i.e., use the target language to formulate a sentence that is intended to recognizably characterize what is visually depicted in the image.

One possible approach to generating such captions is to generate them as such in English, and have them translated (automatically, semi-automatically, or manually) into all the other languages.
However, this approach results in an English-language bias, as well as other problems that have been already identified in the literature.
For instance, translations are often less fluent compared to natural target sentences, due to word order and lexical choices influenced by the source language.
The impact of this phenomenon on metrics and modeling has recently received increased attention in the evaluation literature \cite{toral-etal-2018-attaining,zhang-toral-2019-effect,freitag2020BLEUMB}, and references created in this style are thought to cause overlap-based metrics to favor model outputs that use such unnatural language. 

We have designed our caption annotation process to achieve two main goals:
(i) produce caption annotations in a \textsc{Visible} relation with respect to the image content, and, strongly, create consistency in the description style across languages;
(ii) be free of translation artefacts.
To achieve this, we use bi-lingual annotators with a requirement to be reading-proficient in English and fluent/native in the target language.
As a preliminary step, we train an image-captioning model on English-annotated data, which results in captions in the \textsc{Visible} style of COCO-CAP~\cite{chen2015microsoft}.

The annotation process proceeds as follows.
Each annotation session is done over batches of $N=15$ images, using the images selected as described in Sec.~\ref{sec:image-selection}.
The first screen shows the $N$ images with their captions in English as generated by the captioning model, and asks the annotators if the captions are \textsc{Excellent}, \textsc{Good}, \textsc{Medium}, \textsc{Bad}, or there is \textsc{Not-Enough-Info}. We refer to this rating scale as the 5-level quality scale in the subsequent text.
We provide the annotators with clear guidelines about what constitutes an \textsc{Excellent} caption, and how to evaluate degradations from that quality.
This step forces the annotators to carefully assess caption quality and it primes them into internalizing the style of the captions without the need for complicated and lengthy annotation instructions.

The second round shows the same $N$ images again, but one image at a time without the English captions, and the annotators are asked to produce descriptive captions in the target language for each image.
In the absence of the English captions, the annotators rely on the internalized caption style, and generate their annotations mostly based on the image content -- with no support from the text modality, other than potentially from memory.
Note, however, that we have designed the system to support $N$ annotations simultaneously, and we have empirically selected the value of $N$ as to be large enough to ``overwrite'' the memory of the annotators with respect to the exact textual formulation of the English captions.
As a result, we observe that the produced annotations are free of translation artefacts: See the example in Fig.~\ref{fig:sample_captions} for Spanish mentioning ``number 42'', and for Thai mentioning ``convertibles''.

We also provide the annotators with an annotation protocol to use when creating the captions, which provides useful guidance in achieving consistent annotations across all the targeted languages.
We provide the annotation guidelines in Appendices~\ref{app:rating_instructions} and~\ref{app:generation_instructions}.
For each language, we annotate all \num{3600} images with captions using replication \num{2}
(two different annotators working independently)\footnote{Due to various issues related to process idiosyncrasies, the exact replication varies slightly under or over \num{2}.}, except Bengali (bn) with replication 1 and Maori (mi) with roughly 1 for 2/3 and 2 for 1/3 of the images, see Table~\ref{table:caption_stats}.

\subsection{Caption Statistics}

\begin{table}[ht!]
\rowcolors{2}{gray!15}{white}
\resizebox{\linewidth}{!}{%
\begin{tabular} {cS[table-format=4.0]cS[table-format=4.0]S[table-format=4.0]S[table-format=4.0]cS[table-format=2.1]S[table-format=2.1]}
\toprule
Lan. & {Num.}  && \multicolumn{3}{c}{\hspace{3mm}Replication} & & {Num.} & {Num.}\\
Id.  & {Cap.}  & &1 & 2 & {3+} && {Words} & Chars \\ 
\midrule

ar & 7367 & & 0 & 3434 & 166 & & 7.7 & 42.2 \\
bn & 3600 & & 3600 & 0 & 0 & & 11.3 & 62.1 \\
cs & 7207 & & 15 & 3573 & 12 & & 6.5 & 39.1 \\
da & 7264 & & 0 & 3542 & 58 & & 8.7 & 48.3 \\
de & 8643 & & 0 & 2240 & 1360 & & 11.2 & 76.5 \\
el & 7204 & & 0 & 3596 & 4 & & 7.7 & 51.4 \\
en & 7200 & & 0 & 3600 & 0 & & 9.4 & 49.5 \\
es & 8614 & & 0 & 2201 & 1399 & & 9.8 & 56.3 \\
fa & 7245 & & 0 & 3555 & 45 & & 12.7 & 59.4 \\
fi & 7127 & & 90 & 3500 & 10 & & 7.5 & 65.2 \\
fil & 7109 & & 91 & 3509 & 0 & & 12.2 & 67.6 \\
fr & 8562 & & 0 & 2253 & 1347 & & 12.3 & 69.6 \\
he & 7200 & & 0 & 3600 & 0 & & 11.9 & 63.6 \\
hi & 8503 & & 0 & 2297 & 1303 & & 13.4 & 59.9 \\
hr & 7280 & & 0 & 3553 & 47 & & 9.0 & 57.8 \\
hu & 7216 & & 0 & 3586 & 14 & & 8.5 & 60.5 \\
id & 7126 & & 74 & 3526 & 0 & & 14.3 & 93.5 \\
it & 8471 & & 0 & 2329 & 1271 & & 12.1 & 71.8 \\
ja & 7185 & & 15 & 3585 & 0 & & 1.0 & 26.0 \\
ko & 7650 & & 15 & 3315 & 270 & & 7.0 & 24.7 \\
mi & 4732 & & 2483 & 1102 & 15 & & 11.7 & 55.5 \\
nl & 8059 & & 0 & 2771 & 829 & & 8.0 & 45.9 \\
no & 7213 & & 0 & 3591 & 9 & & 9.6 & 54.3 \\
pl & 7141 & & 59 & 3541 & 0 & & 8.3 & 57.6 \\
pt & 7243 & & 0 & 3562 & 38 & & 10.8 & 61.7 \\
quz & 7200 & & 0 & 3600 & 0 & & 5.0 & 38.6 \\
ro & 7123 & & 77 & 3523 & 0 & & 15.6 & 88.4 \\
ru & 7200 & & 0 & 3600 & 0 & & 9.9 & 66.3 \\
sv & 7273 & & 1 & 3536 & 63 & & 8.1 & 46.7 \\
sw & 7046 & & 154 & 3446 & 0 & & 10.7 & 63.0 \\
te & 7200 & & 0 & 3600 & 0 & & 7.1 & 47.4 \\
th & 7200 & & 0 & 3600 & 0 & & 1.2 & 47.9 \\
tr & 7233 & & 15 & 3538 & 47 & & 9.4 & 63.4 \\
uk & 7215 & & 0 & 3585 & 15 & & 10.0 & 65.7 \\
vi & 7350 & & 0 & 3450 & 150 & & 18.0 & 79.3 \\
zh & 7174 & & 60 & 3508 & 32 & & 1.0 & 23.0 \\
\bottomrule
\end{tabular}}
\caption{\textbf{Caption statistics}: A total of \numcaptions{} captions across \num{36} languages.
We provide the replication stats per language, as well as average number of words (where applicable) and characters.
}
\label{table:caption_stats}
\end{table}
In this section, we take a look at the the basic statistics of the captions in the dataset.
Table \ref{table:caption_stats} provides detailed caption statistics, including the number of captions per image and the average number of words and characters per caption.
There are a total of \numcaptions{} captions across \num{36} languages, each image having in the vast majority of cases at least \num{2} captions per language.

For languages with natural space tokenization, the number of words per caption can be as low as 5 or 6 for some agglutinative languages like Cusco Quechua (quz) and Czech (cs), and as high as \num{18} for an analytic language like Vietnamese (vi).
The number of characters per caption also varies drastically -- from mid-20s for Korean (ko) to mid-90s for Indonesian (id) -- depending on the alphabet and the script of the language.

\begin{table}[ht!]
\centering
\rowcolors{2}{white}{gray!15}
\resizebox{\linewidth}{!}{%
\begin{tabular} {llS[table-format=3.1]S[table-format=3.1]S[table-format=1.1]}
\toprule
\hspace{2mm}Language & Id \hspace{5mm} & \textsc{\%Good+} & \textsc{\%Med+} & \textsc{\%Bad} \\ \midrule
\hspace{2mm}Arabic & ar & 97.5 & 99.3 & 0.7 \\
\hspace{2mm}Bengali & bn & 100.0 & 100.0 & 0.0 \\
\hspace{2mm}Czech & cs & 96.8 & 99.0 & 1.0 \\
\hspace{2mm}Danish & da & 94.0 & 99.2 & 0.8 \\
\hspace{2mm}German  & de & 98.2 & 99.3 & 0.7 \\
\hspace{2mm}Greek & el & 77.3 & 96.0 & 3.7  \\
\hspace{2mm}English & en & 96.5 & 100.0 & 0.0 \\
\hspace{2mm}Spanish  & es & 97.0 & 98.3 & 1.7 \\
\hspace{2mm}Farsi & fa & 94.0 & 99.3 & 0.7 \\
\hspace{2mm}Finnish & fi & 91.5 & 98.8 & 1.2 \\
\hspace{2mm}Filipino & fil & 79.7 & 95.3 & 4.5 \\
\hspace{2mm}French  & fr & 92.7 & 99.2 & 0.8 \\
\hspace{2mm}Hebrew & he & 82.7 & 96.7 & 3.0 \\
\hspace{2mm}Hindi  & hi & 92.7 & 98.7 & 1.3 \\
\hspace{2mm}Croatian & hr & 80.7 & 98.2 & 1.8 \\
\hspace{2mm}Hungarian & hu & 91.3 & 94.8 & 5.0 \\
\hspace{2mm}Indonesian & id & 90.7 & 98.5 & 1.5 \\
\hspace{2mm}Italian  & it & 88.8 & 97.7 & 2.3 \\
\hspace{2mm}Japanese & ja & 84.3 & 96.3 & 3.5 \\
\hspace{2mm}Korean & ko & 85.2 & 99.5 & 0.3 \\
\hspace{2mm}Maori & mi & 93.5 & 98.8 & 1.2  \\
\hspace{2mm}Dutch & nl & 92.8 & 98.7 & 1.3 \\
\hspace{2mm}Norwegian & no & 87.7 & 96.7 & 3.3 \\
\hspace{2mm}Polish & pl & 92.2 & 97.3 & 2.7 \\
\hspace{2mm}Portuguese & pt & 87.8 & 99.5 & 0.3 \\
\hspace{2mm}Cusco Quechua & quz & 83.8 & 98.3 & 1.7 \\
\hspace{2mm}Romanian & ro & 90.2 & 98.3 & 1.7 \\
\hspace{2mm}Russian & ru & 93.8 & 99.5 & 0.3 \\
\hspace{2mm}Swedish & sv & 92.0 & 99.2 & 0.8 \\
\hspace{2mm}Swahili & sw & 70.0 & 98.7 & 1.3 \\
\hspace{2mm}Telugu & te & 98.7 & 99.8 & 0.2 \\
\hspace{2mm}Thai & th & 95.2 & 99.2 & 0.8 \\
\hspace{2mm}Turkish & tr & 97.8 & 98.0 & 1.2 \\
\hspace{2mm}Ukrainian & uk & 91.2 & 99.2 & 0.8 \\
\hspace{2mm}Vietnamese & vi & 94.3 & 97.8 & 2.0 \\
\hspace{2mm}Chinese-Simpl. & zh & 90.2 & 97.8 & 2.2 \\
\bottomrule
\end{tabular}}
\caption{\textbf{Caption quality statistics} for the 36 languages.
We use the median of three ratings as the aggregated rating for an image-caption pair.\vspace{-2mm}}
\label{table:caption_quality}
\end{table}

\subsection{Caption Quality}
In this section, we describe the process for ensuring the creation of high quality annotations, and present quality statistics of the annotations produced. 

In order to ensure quality, the annotation process is initially started with pilot runs on \num{150} images. The caption ratings are spot checked by the authors to verify that the raters have a good understanding of the rating scale. Further, the generated captions go through a verification round where they are rated by the human annotators on the 5-level quality scale described in Sec.\ref{sec:caption-annotation}. If the annotations are below the desired quality, we clarify the guidelines and add more examples to provide feedback to the human annotators and then conduct another pilot. This process is repeated until very few low-quality captions are being produced\footnote{We started the process with a set of six languages and 4-5 pilots were needed per language. For subsequent languages, only 1-2 pilots were needed because of these clarifications and examples added to the guidelines.}.
After this, for every language, we run the main annotation and finally a verification round where we select one caption for \num{600} randomly selected images and have the annotator pool (per language) rate them on the 5-level quality scale mentioned in Sec.~\ref{sec:caption-annotation}.
The quality scores are presented in Table~\ref{table:caption_quality}.

\subsection{Annotator Details}
\label{sec:raters}
We use an in-house annotation platform with professional (paid) annotators and quality assurance.
Annotators are chosen to be native in the target language whenever possible, and fluent otherwise (for low-resource languages, they are usually linguists that have advanced-level knowledge of that language).
All annotators are required to be proficient in English since the instructions and guidelines are given in English.

\section{Model Comparison using \crossmodal{}}


\begin{table*}[ht!]
\centering
\rowcolors{2}{white}{gray!15}
\resizebox{\linewidth}{!}{%
\begin{tabular}{lll} 
 \toprule
 \textbf{Model Name}\hspace{20mm} & \textbf{Details} & \textbf{Parameters} \\
 \midrule
 \baseBcccoco & mT5-base + ViT-B/16 model pretrained on CC3M-35L and finetuned on COCO-35L & {\it lr=\num{3e-4}, cp=10k} \\
 \baseBcoco & mT5-base + ViT-B/16 model trained on COCO-35L & {\it lr=\num{1e-4}, cp=10k}\\ 
 \basegcoco & mT5-base + ViT-g/14 model trained on COCO-35L & {\it lr=\num{1e-4}, cp=10k}\\  
 \largegcoco & mT5-large + ViT-g/14 model trained on COCO-35L & {\it lr=\num{1e-4}, cp=10k}\\ 
 \bottomrule
\end{tabular}}
\caption{\textbf{Model details} for all model variants used in our experiments: \textit{lr} denotes the learning rate; \textit{cp} denotes the number of steps in the constant period where the learning rate is constant.}
\label{table:models}
\end{table*}

\begin{table}[ht!]
\centering
\resizebox{0.8\linewidth}{!}{%
\rowcolors{3}{gray!15}{white}
\setlength{\tabcolsep}{12pt}
\begin{tabular}{cccS[table-format=1.3]S[table-format=1.3]} 
 \toprule
   \bf {\multirow{2}{*}{\bf Model}} &  \bf {\multirow{2}{*}{\bf Lang.}}  & {\bf CIDEr} & {\bf CIDEr} \\  
   & &  {\bf \crossmodal} & {\bf \cocodev} \\
 \midrule
\baseBcccoco{} & en & 0.584 & 0.980 \\
 \baseBcoco{} & en & 0.297 & 0.856 \\
 \basegcoco{} & en & 0.337 & 0.851 \\
 \largegcoco{} & en & 0.343 & 0.875 \\
\midrule
 \baseBcccoco{} & es & 0.425 & 0.962 \\
 \baseBcoco{} & es & 0.194 & 0.844 \\
 \basegcoco{} & es & 0.232 & 0.835 \\
 \largegcoco{} & es & 0.220 & 0.859 \\
\midrule
 \baseBcccoco{} & hi & 0.197 & 0.759 \\
 \baseBcoco{} & hi & 0.098 & 0.671 \\
 \basegcoco{} & hi & 0.112 & 0.718 \\
 \largegcoco{} & hi & 0.111 & 0.624 \\
\midrule
 \baseBcccoco{} & zh & 0.202 & 0.748 \\
 \baseBcoco{} & zh & 0.087 & 0.659 \\
 \basegcoco{} & zh & 0.110 & 0.695 \\
 \largegcoco{} & zh & 0.099 & 0.656 \\
\bottomrule
\end{tabular}}
\caption{\textbf{CIDEr on \crossmodal\ and \cocodev} for the models over the four languages \lfour{} (\cocodev{} computed using machine-translated references). Tables~\ref{table:raw_cider_baseBcccoco}-\ref{table:raw_cider_basebcoco} in the appendix show all the CIDEr values for all the models.
}
\label{table:raw_cider}
\end{table}

In this section, we detail our experiments for comparing several models using human evaluations, and also using \crossmodal{} annotations as gold\footnote{We use ``gold'' here to refer to human-level quality; references created via means like automatic translation are refered to as ``silver'' quality.} references for automated metrics.

For model comparison, we train several multilingual image captioning models with different sizes over different datasets, and compare them on \crossmodal{}. As our main result, we show a high level of correlation between model rankings based on human-evaluation scores and the scores obtained using CIDEr~\cite{cider} with \crossmodal{} annotations as gold references.

\subsection{Datasets}
We build two multilingual datasets for training, CC3M-35L and COCO-35L, by translating Conceptual Captions 3M~\cite{sharma2018conceptual} and COCO Captions~\cite{chen2015microsoft} to the other 34 languages using Google's machine translation API\footnote{\url{https://cloud.google.com/translate}}. The remaining language, Cusco Quechua (quz), is not supported by the API\footnote{ Although the translate.google.com website has recently added support, the API does not support it as of this writing.}. We use the standard train and validation splits for CC3M\footnote{CC3M-train: \num{3318333} image-caption pairs. CC3M-validation: \num{15840} image-caption pairs.}. For COCO, we use the \textit{Karpathy split}~\cite{karpathy_coco:2014}\footnote{Train: \num{113287} images. Validation: \num{5000} images. Each image has 5 reference captions.}.

\subsection{Models}
In this section we detail the model architecture we used for the experiments.
\begin{figure}
\begin{minipage}[t]{0.47\linewidth}
\centering
\includegraphics[height=4.2cm]{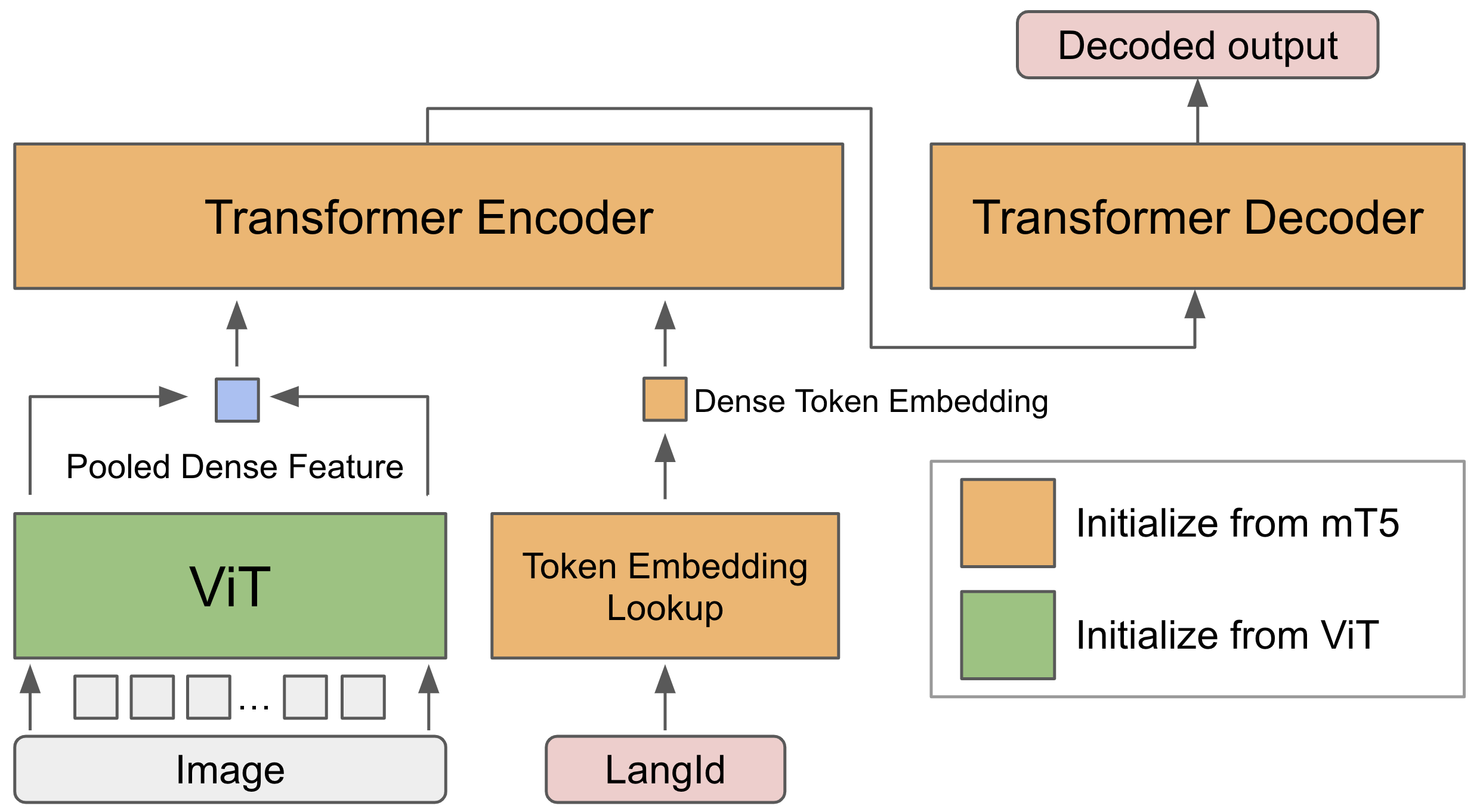}
\end{minipage}
\vspace{-2mm}
\caption{The architecture for the family of multilingual image captioning models used in the experiments.}
\label{fig:model}
\end{figure}

Our Transformer-based \cite{transformers} model architecture for image captioning is shown in Figure \ref{fig:model}. On the vision side, each input image is modeled by a Vision Transformer (ViT) \cite{vit:2020,scaling_vit:2021}. The visual features produced by ViT for every patch of the image are pooled into a single dense feature vector. On the text side, a Language Identifier (LangId) string is used to specify the language. The LangId string is tokenized and embedded into dense token embeddings, which are merged with the dense visual embeddings as the input to a multi-layer Transformer Image and Text Encoder, followed by a multi-layer Transformer Image and Text Decoder to generate the predicted captions. 

We take advantage of existing pretrained models to initialize different parts of our model: ViT~\cite{scaling_vit:2021} (green in Fig.~\ref{fig:model}) and mT5~\cite{xue-etal-2021-mt5} (orange in Fig.~\ref{fig:model}).  We consider different model sizes: mT5-base, mT-large, ViT-B/16, and ViT-g/14, where 16 and 14 are the corresponding patch sizes.
We choose three combinations resulting in three different model architectures: mT5-base + ViT-B/16, mT5-base + ViT-g/14 and mT5-large + ViT-g/14.

We train these three models on COCO-35L. In addition, we consider a fourth model based on mT5-base + ViT-B/16 and trained on CC3M-35L. The models are trained on a 4x4x4 TPU-v4 architecture using an Adafactor~\cite{shazeer2018adafactor} optimizer with a constant learning rate period between \{\num{1}k, \num{10}k\} steps, followed by a reversed square-root decay with the number of steps. The batch size is \num{2048} in all the experiments. The initial learning rate is between \{1e-4, 3e-4\}. We use the same vocabulary (size 250k) as mT5~\cite{xue-etal-2021-mt5}. The model trained with CC3M-35L is subsequently finetuned on COCO-35L with constant learning rate 3e-5 for 1 epoch.

Table~\ref{table:models} describes the best hyperparameters we found for different training setups and used in our quantitative experiments.
In terms of model sizes, mT5-base has about 680 million parameters, mT5-large about 1230 million, the ViT-B/16 86 million, and ViT-g/14 1011 million parameters.
Together, all the experiments took around \num{5000} TPU hours to train.

\subsection{Human Evaluation}
\label{sec:human_eval}
In this section, we detail the process used for human evaluations comparing the performance of two models.

Our main goal in creating \crossmodal{} is to automate the evaluation of massively multilingual image captioning models, by eliminating expensive and time-consuming human evaluations.
Our results indicate that they can be substituted by using the \crossmodal{} annotations as gold references for automated metrics such as CIDEr~\cite{cider}.
To quantify the correlation between the two methods, we train four different models (Tab.~\ref{table:models}) and conduct side-by-side human evaluations using the outputs of these models in several languages.
We observe strong correlations (Sec.~\ref{sec:results}) between the human evaluations and the CIDEr scores using the \crossmodal{} references.

Specifically, we use a randomly selected subset of 600 images from \crossmodal{} for human evaluations, which we call \xmsix{}. 
Image captions generated by a given pairing of models 
($m_1$ vs $m_2$, where $m_1$ is considered as the base condition and $m_2$ as the test condition)
are compared and rated side-by-side, using a similar pool of annotators as described in Sec.~\ref{sec:raters}.
Each side-by-side pair (shown in a random per-example left-vs-right order) is rated using a 7-point scale: \textsc{\small Much-Better}, \textsc{\small Better}, \textsc{\small Slightly-Better}, \textsc{\small Similar}, \textsc{\small Slightly-Worse}, \textsc{\small Worse}, \textsc{\small Much-Worse}, with a replication factor of 3 (three annotators rate each pair).
We denote by \wins{} the percentage of images where the majority of raters (i.e. 2 out of 3) mark $m_2$'s captions as better, and by \losses{} the percentage of images where the majority of raters mark $m_2$'s captions as worse.
We then define the overall side-by-side gain of $m_2$ over $m_1$ as \sxsgain{}  = \wins{} - \losses.

Conducting the full set of six side-by-side evaluations for each pair of models over the 35 languages would require 210 human evaluation sessions. This is prohibitively expensive and time consuming. Thus, we conduct the full set of six side-by-side evaluations of the pairs of models, on a core set of four languages called \lfour{}\footnote{\langsfour{}}. We call this set of 24 evaluation sessions \ofour{}. Furthermore, we also conduct a sparser set of side-by-side evaluations over languages where the CIDEr differences on \crossmodal{} and on \cocodev{}\footnote{COCO validation split with machine-translated references} indicate disagreement or ambiguity (e.g., opposite sign of the CIDEr differences, and/or small CIDEr differences); this gives us a set of 28 languages called \leval{}\footnote{\langseval{}}. We call the resulting set of 41 evaluation sessions \oeval{}. The set of all evaluations is called \oall{} $=$\ofour{} + \oeval{}, which are conducted over the languages \lall{}  $=$ \lfour{} + \leval{}.

The choice of which model is called $m_1$ and which model is called $m_2$ is arbitrary in the side-by-side evaluations, since we randomly flip left vs right before presenting the captions to the raters. Hence a single side-by-side evaluation gives two points for the correlation calculations: one with the $m_1$ and $m_2$ assigned as per the actual evaluation conducted, and one more with the $m_1$ and $m_2$ assignment flipped and the \sxsgain{} sign flipped correspondingly.

\begin{table}[ht!]
\centering
\resizebox{\linewidth}{!}{%
\rowcolors{3}{gray!15}{white}
\begin{tabular}{cccS[table-format=3.1]S[table-format=1.2]S[table-format=1.2]S[table-format=1.2]} 
 \toprule
   \bf {\multirow{2}{*}{$\mathbf m_2$}} & \bf {\multirow{2}{*}{$\mathbf m_1$}} & {\multirow{2}{*}{\bf L.}} & {\multirow{2}{*}{\bf \sxsgain{}}}  & {\bf $\Delta$CIDEr} & {\bf $\Delta$CIDEr} & {\bf $\Delta$CIDEr} \\  
   &        &    &  & {\bf \xmsix{}} & {\bf \crossmodal} & {\bf \cocodev}\\
 \midrule
 \baseBcccoco{} & \baseBcoco{} & en & -38.9 & -0.277 & -0.287 & -0.124 \\
 \baseBcccoco{} & \basegcoco{} & en & -21.5 & -0.230 & -0.247 & -0.129 \\
 \baseBcccoco{} & \largegcoco{} & en & -34.8 & -0.246 & -0.240 & -0.105 \\
 \basegcoco{} & \largegcoco{} & en & -3.9 & -0.016 & 0.007 & 0.024 \\
 \basegcoco{} & \baseBcoco{} & en & -14.0 & -0.047 & -0.039 & 0.005 \\
 \largegcoco{} & \baseBcoco{} & en & -10.3 & -0.031 & -0.046 & -0.018 \\
 \basegcoco{} & \largegcoco{} & es & -1.3 & 0.002 & -0.012 & 0.024 \\
 \basegcoco{} & \baseBcoco{} & es & -8.4 & -0.044 & -0.037 & 0.008 \\
 \largegcoco{} & \baseBcoco{} & es & -4.4 & -0.045 & -0.026 & -0.016 \\
 \baseBcccoco{} & \largegcoco{} & es & -29.6 & -0.201 & -0.205 & -0.103 \\
 \baseBcccoco{} & \basegcoco{} & es & -28.8 & -0.203 & -0.193 & -0.127 \\
 \baseBcccoco{} & \baseBcoco{} & es & -36.5 & -0.246 & -0.231 & -0.118 \\
 \basegcoco{} & \largegcoco{} & hi & -3.0 & -0.001 & -0.001 & -0.094 \\
 \baseBcccoco{} & \basegcoco{} & hi & -29.3 & -0.095 & -0.084 & -0.040 \\
 \largegcoco{} & \baseBcoco{} & hi & -2.0 & -0.012 & -0.013 & 0.047 \\
 \baseBcccoco{} & \baseBcoco{} & hi & -36.5 & -0.108 & -0.099 & -0.088 \\
 \basegcoco{} & \baseBcoco{} & hi & -5.2 & -0.013 & -0.015 & -0.047 \\
 \baseBcccoco{} & \largegcoco{} & hi & -32.3 & -0.096 & -0.086 & -0.135 \\
 \basegcoco{} & \largegcoco{} & zh & -8.9 & -0.018 & -0.012 & -0.039 \\
 \baseBcccoco{} & \largegcoco{} & zh & -30.0 & -0.104 & -0.103 & -0.092 \\
 \baseBcccoco{} & \basegcoco{} & zh & -30.1 & -0.086 & -0.092 & -0.053 \\
 \baseBcccoco{} & \baseBcoco{} & zh & -41.2 & -0.102 & -0.115 & -0.089 \\
 \basegcoco{} & \baseBcoco{} & zh & -16.0 & -0.016 & -0.023 & -0.036 \\
 \largegcoco{} & \baseBcoco{} & zh & -11.8 & 0.002 & -0.011 & 0.003 \\
\bottomrule
\end{tabular}}
\caption{\textbf{Model comparisons over \lfour{} languages} ($m_2$ vs $m_1$). \textit{L} denotes the target language; $\Delta$CIDEr \xmsix{} is CIDEr($m_2$)-CIDEr($m_1$) on the \xmsix{} dataset, $\Delta$CIDEr \crossmodal{} on the \crossmodal{} dataset, and $\Delta$CIDEr \cocodev{} on the COCO validation split with machine-translated references.
Table~\ref{table:detailed_model_comparison} in the appendix shows model comparisons over the \leval{} languages.}
\label{table:model_comparison}
\end{table}

\begin{table}[t!]
\centering
\resizebox{\linewidth}{!}{%
\rowcolors{3}{gray!15}{white}
\begin{tabular}{cccS[table-format=1.2]S[table-format=1.2]S[table-format=1.2]} 
 \toprule
 \textbf{Correlation} & {\multirow{2}{*}{\bf Lang.}} & {\multirow{2}{*}{\bf N}} &{\textbf{$\Delta$CIDEr}} & {\textbf{$\Delta$CIDEr}} & {\textbf{$\Delta$CIDEr}} \\  
 \textbf{Coefficient} & & & {\textbf{\xmsix}} & {\textbf{\crossmodal}} & {\textbf{\cocodev}} \\
 \midrule
 Pearson & \lall{} & 130 & 0.88 & 0.88 & 0.68 \\
 Spearman & \lall{} & 130 & 0.87 & 0.92 & 0.30 \\
 Kendall & \lall{} & 130 & 0.69 & 0.76 & 0.21 \\
 \midrule
 Pearson & \leval{} & 82 & 0.72 & 0.84 & -0.44 \\
 Spearman & \leval{} & 82 & 0.76 & 0.84 & -0.52 \\
 Kendall & \leval{} & 82 & 0.54 & 0.65 & -0.32 \\
 \midrule
 Pearson & \lfour{} & 48 & 0.90 & 0.90 & 0.89 \\
 Spearman & \lfour{} & 48 & 0.95 & 0.96 & 0.86 \\
 Kendall & \lfour{} & 48 & 0.80 & 0.81 & 0.67 \\
\bottomrule
\end{tabular}}

\caption{\textbf{Correlations} between side-by-side human evaluations (\sxsgain{}) and CIDEr difference on \xmsix{}, \crossmodal{} and the translated COCO validation set. Here N represents the number of points used to compute the correlation coefficient. As noted in Sec.~\ref{sec:human_eval}, each evaluation gives us two points for the correlation calculation. \vspace{-2mm}}
\label{table:corr}
\end{table}

\subsection{Results}
\label{sec:results}
We present results that show that it is feasible to use the \crossmodal{} annotations as gold references with automated metrics such as CIDEr to compare models in lieu of human evaluations, and that this option is superior to using silver references created via automated translation.

Table~\ref{table:model_comparison} presents the results for the \ofour{} set of evaluations on \xmsix{} on the \lfour{} languages, while Table~\ref{table:detailed_model_comparison} in the appendix shows the results on the \leval{} languages.  The reference for the relative strength of each pairing is given by \sxsgain{}, with positive numbers indicating the superiority of $m_2$, and negative numbers indicating a superiority of $m_1$. As can be seen from the table, the model comparisons span a range of model differences, from low \sxsgain{} to high \sxsgain{}. 
$\Delta$CIDEr \xmsix and $\Delta$CIDEr \crossmodal{} capture similar information, except these numbers are based on CIDEr scores using as references \xmsix{} and \crossmodal{}, respectively, while $\Delta$CIDEr \cocodev{} is based on machine-translated references from the validation split of COCO.

We use the results from Table~\ref{table:model_comparison} (and Table~\ref{table:detailed_model_comparison}) to compute the correlation between human judgements of the relative quality of the captioning models and the ability of the CIDEr\footnote{We use the reference implementation with default parameters: \href{https://github.com/vrama91/cider}{github.com/vrama91/cider}. We remove punctuation and lowercase the captions and references before computing automated metrics.} metric -- or, rather, of the underlying references used by the metric -- to perform an equivalent task.
Table~\ref{table:corr} presents the correlation results using three correlation metrics: Pearson, Spearman, and Kendall. The first section shows the correlations over all the side-by-side evaluations (i.e. \ofour{} and \oeval{}); These cover the \lfour{} and the \leval{} languages. The second section shows the correlations for the \oeval{} covering the \leval{} languages. The third section shows the correlations for the \ofour{} evaluations covering the \lfour{} languages. 

We observe that $\Delta$CIDEr \crossmodal{} is highly correlated with human judgement according to all the correlation metrics \cite{bonnett:00}, over all the evaluations \oall{}, over the \ofour{} evaluations, and also the \oeval{} evaluations. Furthermore, for the \oeval{} evaluations, where most of the instances have opposite signs for $\Delta$CIDEr \cocodev{} and $\Delta$CIDEr \crossmodal{}, we find that the former is strongly anti-correlated with the human evaluation results while the latter is highly correlated with the human evaluation results.
Overall, these results indicate that: (i) we can reliably substitute $\Delta$CIDEr \crossmodal{} for human evaluations on \xmsix{} when comparing models similar to the ones we used; (ii) the gold \crossmodal{} references are preferable over the silver references obtained from translating COCO captions, in terms of approximating the judgements of the human evaluators\footnote{However, it is unclear whether machine translated references for one particular language in \crossmodal{} translated to all others, are worse than using the human generated references. In particular, we studied the correlations of CIDEr computed using \xmmt{en} (i.e. the \crossmodal{} English references, machine translated to all the other languages), with the human evaluations. We found that even though the translations have artifacts and disfluencies, CIDEr differences calculated using them show comparable correlations with human judgement observations. We also studied such correlations for machine translated references from German, Greek, Hebrew, Hungarian and Swahili. We found that the correlations are similar and sometimes even a bit higher than using the human generated references. We believe this happens because the rater guidelines weigh informativeness over fluency and the CIDEr metric is also not as sensitive to fluency. Further work is needed to understand the use of translated references as compared to human generated references. We believe that using the human generated references along with the set of machine translated references from all the other languages may provide even stronger correlations and will show greater diversity in the coverage of the image constituents. }.

Based on the results from Table~\ref{table:corr}, we recommend the use of the \crossmodal{} references as a means to achieve high-quality automatic comparisons between multilingual image captioning models. We have provided the CIDEr scores for \crossmodal{} in 35 languages for all the models, in  Tables~\ref{table:raw_cider_baseBcccoco}-\ref{table:raw_cider_basebcoco} in the Appendix. These can be used as baselines in future work.

\section{Conclusions}
We introduce the \crossmodal{} dataset as a benchmark for evaluating the performance of multilingual image captioning models.
The images in the dataset are geographically diverse, covering all inhabited continents and a large fraction of the world population.
We believe this benchmark has the potential to positively impact both the research and the applications of this technology, and enable (among other things) better accessibility for visually-impaired users across the world, including speakers of low-resource languages.

The main appeal of this benchmark is that it alleviates the need for extensive human evaluation, which is difficult to achieve across multiple languages and hinders direct comparison between different research ideas and results.
We show significant improvements in correlation with human judgements when using the \crossmodal{} dataset as references for automatic metrics, and therefore hope that the adoption of this dataset as a standard benchmark will facilitate faster progress and better comparisons among competing ideas. 

Our empirical observations are primarily on the full set of side-by-side comparisons over English and three other languages (Spanish, Hindi, Chinese).
Due to the similarity in the data collection and the quality control process, we expect similar results to hold for all the other languages as well;
we validated this expectation with additional empirical observations covering an additional 28 languages.

\section{Limitations}


Due to the high volume of work required and the cost associated with it, we have only targeted 36 languages for our annotation effort; while this number is significantly higher than what is available with previous annotations, it still falls short of including many other languages spoken and written around the world. Additionally, since the \lthirty{} languages were selected based on their internet presence, one unintended consequence is that the dataset over-represents European languages. While this is somewhat mitigated by including the \lfive{} low resource languages, building and sharing this dataset can have the unintended effect of perpetuating the issue where computational linguistics and AI work is often unintentionally Eurocentric. 

Due to the cost and logistical constraints, we have sampled only 100 images for each of the targeted languages, which limits the amount of natural and cultural phenomena that these images capture.
While the resulting 3600 images have significantly more variety compared to previous datasets, it may still fall short of including important aspects of natural and cultural life from around the globe. Further, there is the possibility of bias in the dataset due to the uneven access to photographic equipment and internet connectivity (For example, several of the images in Fig. \ref{fig:additinal_examples} seem be shared by people with non-native names in the context of the locales. Thus, these images may have been taken by tourists rather than natives. Further exploration into this aspect of the dataset is important as well).

Another limitation is around the absence of translation artifacts in the annotations. We primarily rely on the caption generation process outlined in Sec. \ref{sec:caption-annotation} and on rater quality controls for avoiding translation artifacts. Further, we have performed spot checks on captions in several languages and have not found indications of translation artifacts. Additionally, we have also compared the translations of annotations from another language such as English with generated annotations and verified that the translations show peculiar artifacts and disfluencies which are not seen in the generated annotations. 

We would also like to emphasize that, while this dataset aims to ameliorate the need for human evaluations for multilingual image captioning, automated evaluation may be less sensitive to small changes, e.g. when comparing highly tuned methods submitted to competitions. This was one of our motivations for comparing models that range from very different (CC+Bg vs Bg/Lg/BB)) to moderately different (BB vs Bg/Lg) and quite similar (Bg vs Lg), and the results from Table \ref{table:model_comparison} show that our approach works well over this range of model differences over \lfour{}. We also stress-tested our approach by focusing the \oeval{} evaluations on cases where $\Delta$CIDEr \crossmodal{} or $\Delta$CIDEr \cocodev{} were quite small or of opposite signs, and the results from Table 7 in the appendix show that $\Delta$CIDEr \crossmodal{} correlated well with human evaluations even for this harder set of evaluations. However, we caution the reader that there will be cases where human judgement will still be needed. Further, automated evaluations may be biased to methods that explicitly optimize the evaluated metric, e.g. via approaches such as Self-Critical Sequence Training\cite{rennie:17}.

We also note that the model outputs and human judgements data used for calculating the correlations would be useful for constructing new automated metrics and validating existing automated metrics for model comparisons. Releasing this data would also allow independent calculation of CIDEr and \sxsgain{} shown in Table \ref{table:model_comparison} and Table \ref{table:detailed_model_comparison}. However, due to the timelines involved and approvals required, we are not able to release this data with the paper. This may hamper the reproducibility of these computations. 

The approach to data collection and annotation of COCO-CAP~\cite{chen2015microsoft} and CC3M~\cite{sharma2018conceptual} upholds rigorous privacy and ethics standards, such as the avoidance of offensive content and exposure of personal identification data.
This significantly mitigates but does not completely eliminate the risks that the captioning models we train would produce such information.
Similarly, the \crossmodal{} dataset mitigates such risks by adopting a defense-in-depth approach: 1) The annotations have been produced in-house and have been quality controlled, while the images used have been vetted to be appropriate for the intended use. 2) Further, the machine translations of the annotations have been scanned with an automated tool to detect personally identifiable information. 3) The machine translations of the annotations have been spot-checked by the authors.

Overall, in spite of the above limitations, we believe that this dataset is a significant step toward ameliorating language and geographic bias, and that it should be used for advancing image captioning research over a wider variety of images and languages.

\section{Acknowledgements}
We would like to thank the anonymous reviewers for providing feedback which led to several improvements such as: 1) A discussion about correlations of human judgement with the machine translations of the \crossmodal{} references; 2) A discussion about the possibility of releasing model outputs and human evaluation data which may help with reproducibility and also help evaluation of existing and new automated metrics over \lall{}. 

\bibliography{anthology,custom}
\bibliographystyle{acl_natbib}

\appendix

\section{Additional Caption Examples}
\label{app:additional_captions}
\begin{figure*}
\vspace{-8mm}
\setlength{\fboxsep}{0pt}%
  \centering
  \begin{minipage}{0.35\linewidth}
  \centering
  \fbox{\includegraphics[width=0.9\linewidth]{figures/06c5f87181c30e00.jpeg}}\\[-1mm]
  {\tiny\it Source: \href{https://www.flickr.com/photos/briansolis/5129089526}{Porsche Museum, Stuttgart} by \href{https://www.flickr.com/people/briansolis/}{Brian Solis}.}
  \end{minipage}\\
  \mbox{\hspace{-6mm}\includegraphics[trim=18mm 65mm 16mm 17mm, clip,width=1\linewidth]{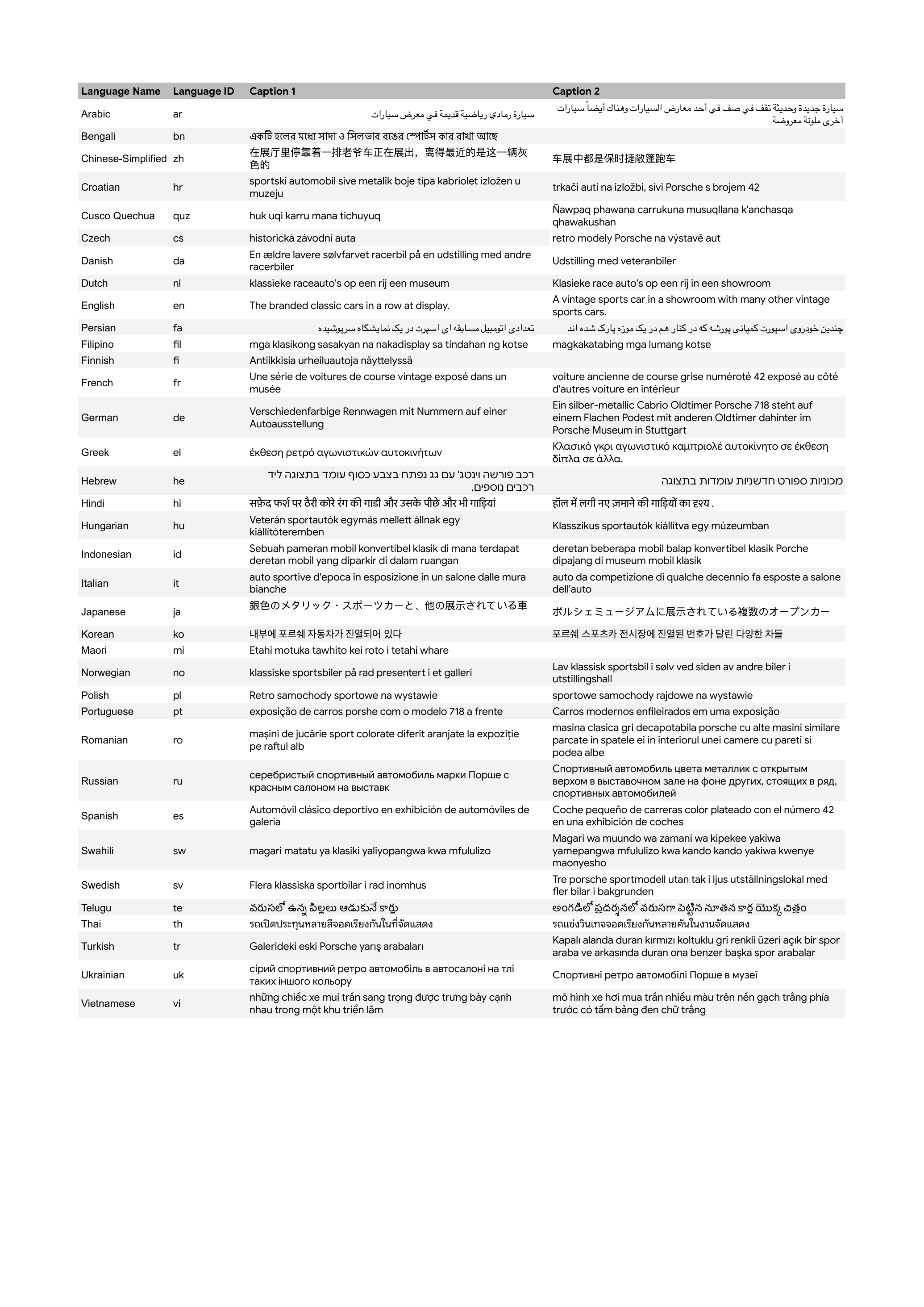}}
  \caption{Example captions in the 36 languages covered in \crossmodal}
  \label{fig:additinal_examples}
\end{figure*}

Figure~\ref{fig:additinal_examples} displays the captions in the 36 languages covered in \crossmodal{} for the same image as in Figure~\ref{fig:sample_captions}.

\section{Instructions for Rating Captions}
\label{app:rating_instructions}
The following instructions are provided to annotators for rating captions:\\[-1mm]

{\small This task involves rating captions. To guide your ratings, imagine that you are describing the image to a visually impaired friend, then consider: how well does the caption describe the image to this friend?

Use the following scale for judging the quality of the captions (for borderline cases, use the lower rating):
\begin{itemize}
    \item{\textsc{Bad}}: The caption has one or more of the following issues:
       a). Caption misses the main topic of the image.
       b). Caption has major grammatical errors (such as being incomplete, words in wrong order, etc). Please ignore capitalization of words and punctuation.
       c). Caption violates the ‘No Hallucination’ rule by mentioning objects, activities, or relationships that are definitely not in the image.
            Note: Apply the ‘No-Hallucination’ rule only when you are certain that an object/activity/relationship is definitely not implied by the image (see the examples below).
   \item{\textsc{Mediocre}}: The caption may capture some objects and activities but misses crucial information (related to activity, important objects/persons in the scene, important modifiers, etc.)
   \item{\textsc{Good}}: The caption explains most of the main objects, activities, and their relationships in the image.
   \item{\textsc{Excellent}}: The caption covers well the whole image, including all the main objects, activities, and their relationships.
   \item{\textsc{Not Enough information}}: Not enough information to evaluate the caption quality. Please try to use one of the four categories above as much as possible. Assume that any missing information is favorable to the caption rather than against it.
\end{itemize}
}

\section{Instructions for Generating Captions}
\label{app:generation_instructions}
The following instructions are provided to annotators for generating captions:\\[-1mm]

{\small 
To guide your caption generation, imagine that you are describing the image to a visually impaired friend. The caption should explain the whole image, including all the main objects, activities, and their relationships. The objects should be named as specifically as practical: For example when describing a young boy in a picture, “young boy” is preferred over “young child”, which in turn is preferred over “person”.

Note: the goal is to generate captions that would be labeled as “Excellent” under the Rating guidelines above, but raters should not copy captions from the first phase. We want the raters to generate the captions on their own.

We outline here a procedure that you should try and follow when writing your image caption. Note that not all these steps may be applicable for all images, but they should give you a pretty good idea of how to organize your caption. We will make use of the first image in the table below (the one with the young girl smiling) 
Note: It is acceptable to make assumptions that are reasonable as long as they don’t contradict the information in the image (eg: in the second image below, we use “families” in captions 1 and 3 because there seems to be a mix of children and adults though it is not perfectly clear. So it is a reasonable assumption to make and nothing in the image contradicts it. However it is also ok to use “people”.)
\begin{enumerate}
\item Identify the most salient objects(s)/person(s) in the image; use the most informative level to refer to something (i.e., “girl” rather than “child” or “person”); in the example image: “girl”
\item Identify the most salient relation between the main objects; example “girl standing in front of the whiteboard”
\item Identify the main activity depicted; in the example image: “smiling” as an activity (note that this can also be an attribute of the girl), or “standing” as an activity
\item Identify the most salient attributes of the main object(s)/person(s)/activity(es); in the example image: “smiling” and “young” as attributes for the girl
\item Identify the background/context/environment in which the scene is placed; in the example image: “classroom”
\item Put everything together from steps 1-5 above; for the example image: “a smiling girl standing in a classroom”, or “a young girl smiling in a classroom”.
\end{enumerate}
}

\section{Detailed Results for Model Comparison}
\begin{table}[ht!]
\centering
\resizebox{\linewidth}{!}{%
\rowcolors{3}{gray!15}{white}
\begin{tabular}{cccS[table-format=3.1]S[table-format=1.2]S[table-format=1.2]S[table-format=1.2]} 
 \toprule
   \bf {\multirow{2}{*}{$\mathbf m_2$}} & \bf {\multirow{2}{*}{$\mathbf m_1$}} & {\multirow{2}{*}{\bf L.}} & {\multirow{2}{*}{\bf \sxsgain{}}}  & {\bf $\Delta$CIDEr} & {\bf $\Delta$CIDEr} & {\bf $\Delta$CIDEr} \\  
   &        &    &  & {\bf \xmsix{}} & {\bf \crossmodal} & {\bf \cocodev}\\
 \midrule
 \largegcoco{} & \baseBcoco{} & ar & -1.2 & -0.003 & -0.003 & 0.055 \\
 \largegcoco{} & \baseBcoco{} & bn & -3.5 & -0.025 & -0.026 & 0.039 \\
 \basegcoco{} & \baseBcoco{} & cs & -5.2 & -0.031 & -0.016 & 0.013 \\
 \largegcoco{} & \baseBcoco{} & cs & -2.7 & -0.012 & 0.002 & 0.029 \\
 \basegcoco{} & \baseBcoco{} & da & -6.9 & -0.021 & -0.029 & 0.048 \\
 \basegcoco{} & \largegcoco{} & da & 2.7 & -0.009 & -0.003 & 0.029 \\
 \largegcoco{} & \baseBcoco{} & da & -13.3 & -0.012 & -0.026 & 0.018 \\
 \basegcoco{} & \baseBcoco{} & de & -12.6 & -0.014 & -0.026 & 0.030 \\
 \basegcoco{} & \largegcoco{} & de & -1.5 & -0.007 & -0.008 & 0.037 \\
 \largegcoco{} & \baseBcoco{} & de & -9.6 & -0.007 & -0.018 & -0.006 \\
 \largegcoco{} & \baseBcoco{} & el & -5.1 & -0.005 & 0.002 & 0.063 \\
 \largegcoco{} & \baseBcoco{} & fa & -11.1 & -0.011 & -0.003 & 0.027 \\
 \largegcoco{} & \baseBcoco{} & fi & -0.3 & -0.007 & -0.006 & 0.008 \\
 \largegcoco{} & \baseBcoco{} & fil & -3.2 & -0.024 & -0.004 & 0.020 \\
 \largegcoco{} & \baseBcoco{} & fr & -2.0 & -0.030 & -0.015 & 0.011 \\
 \basegcoco{} & \baseBcoco{} & fr & -3.0 & -0.022 & -0.024 & 0.001 \\
 \largegcoco{} & \baseBcoco{} & he & 1.7 & 0.005 & 0.001 & 0.025 \\
 \largegcoco{} & \baseBcoco{} & hr & -4.5 & -0.007 & 0.002 & 0.030 \\
 \basegcoco{} & \baseBcoco{} & hr & -8.4 & -0.019 & -0.023 & 0.014 \\
 \largegcoco{} & \baseBcoco{} & hu & -4.7 & -0.005 & -0.009 & 0.027 \\
 \largegcoco{} & \baseBcoco{} & id & -4.7 & -0.031 & -0.018 & 0.004 \\
 \largegcoco{} & \baseBcoco{} & it & -6.2 & -0.011 & -0.015 & 0.010 \\
 \largegcoco{} & \baseBcoco{} & ja & -10.8 & -0.013 & -0.021 & -0.006 \\
 \largegcoco{} & \baseBcoco{} & ko & -2.0 & -0.025 & -0.018 & 0.045 \\
 \basegcoco{} & \baseBcoco{} & nl & -7.7 & 0.003 & -0.016 & 0.009 \\
 \basegcoco{} & \largegcoco{} & nl & 0.2 & 0.001 & -0.003 & 0.007 \\
 \largegcoco{} & \baseBcoco{} & nl & -6.2 & 0.002 & -0.013 & 0.002 \\
 \largegcoco{} & \baseBcoco{} & no & -10.3 & -0.043 & -0.033 & 0.007 \\
 \basegcoco{} & \baseBcoco{} & pl & -3.5 & -0.022 & -0.023 & 0.015 \\
 \largegcoco{} & \baseBcoco{} & pl & -4.4 & 0.002 & -0.007 & 0.019 \\
 \largegcoco{} & \baseBcoco{} & pt & -10.8 & -0.027 & -0.026 & -0.011 \\
 \largegcoco{} & \baseBcoco{} & ro & -5.6 & -0.027 & -0.017 & -0.001 \\
 \largegcoco{} & \baseBcoco{} & sv & -8.1 & -0.028 & -0.035 & 0.003 \\
 \basegcoco{} & \largegcoco{} & sv & -5.4 & 0.004 & 0.004 & 0.031 \\
 \basegcoco{} & \baseBcoco{} & sv & -9.3 & -0.024 & -0.032 & 0.035 \\
 \largegcoco{} & \baseBcoco{} & sw & -2.2 & -0.016 & 0.007 & 0.040 \\
 \largegcoco{} & \baseBcoco{} & te & -1.3 & 0.008 & -0.002 & 0.037 \\
 \largegcoco{} & \baseBcoco{} & th & -6.7 & -0.031 & -0.016 & 0.028 \\
 \largegcoco{} & \baseBcoco{} & tr & -5.1 & -0.026 & -0.016 & 0.029 \\
 \basegcoco{} & \baseBcoco{} & vi & -4.5 & 0.000 & 0.000 & 0.058 \\
 \largegcoco{} & \baseBcoco{} & vi & 3.2 & 0.015 & 0.008 & 0.082 \\
\bottomrule
\end{tabular}}
\caption{\textbf{Model comparison over the \leval{} languages} ($m_2$ vs $m_1$). \textit{L} denotes the target language; $\Delta$CIDEr \xmsix{} is CIDEr($m_2$)-CIDEr($m_1$) on the \xmsix{} dataset, $\Delta$CIDEr \crossmodal{} on the \crossmodal{} dataset, and $\Delta$CIDEr \cocodev{} on the COCO validation split with machine-translated references.}
\label{table:detailed_model_comparison}
\end{table}

\begin{table}[ht!]
\centering
\resizebox{0.6\linewidth}{!}{%
\rowcolors{3}{gray!15}{white}
\setlength{\tabcolsep}{12pt}
\begin{tabular}{cccS[table-format=1.3]S[table-format=1.3]} 
 \toprule
   \bf \bf {\multirow{2}{*}{\bf Lang.}}& {\bf CIDEr} & {\bf CIDEr} \\  
& {\bf \crossmodal} & {\bf \cocodev} \\
 \midrule
 ar & 0.227 & 0.649 \\
 bn & 0.200 & 0.682 \\
 cs & 0.313 & 0.575 \\
 da & 0.329 & 0.877 \\
 de & 0.224 & 0.735 \\
 el & 0.199 & 0.830 \\
 en & 0.584 & 0.980 \\
 es & 0.425 & 0.962 \\
 fa & 0.311 & 0.898 \\
 fi & 0.177 & 0.487 \\
 fil & 0.353 & 1.007 \\
 fr & 0.410 & 0.957 \\
 he & 0.230 & 0.650 \\
 hi & 0.197 & 0.759 \\
 hr & 0.224 & 0.607 \\
 hu & 0.175 & 0.551 \\
 id & 0.307 & 1.088 \\
 it & 0.321 & 0.902 \\
 ja & 0.254 & 0.963 \\
 ko & 0.288 & 0.862 \\
 mi & 0.405 & 1.175 \\
 nl & 0.441 & 0.796 \\
 no & 0.385 & 0.856 \\
 pl & 0.236 & 0.578 \\
 pt & 0.380 & 0.964 \\
 ro & 0.188 & 0.832 \\
 ru & 0.194 & 0.675 \\
 sv & 0.370 & 0.848 \\
 sw & 0.319 & 0.796 \\
 te & 0.196 & 0.520 \\
 th & 0.418 & 0.929 \\
 tr & 0.232 & 0.668 \\
 uk & 0.189 & 0.653 \\
 vi & 0.336 & 1.150 \\
 zh & 0.202 & 0.748 \\
\bottomrule
\end{tabular}}
\caption{\textbf{CIDEr on \crossmodal{} and \cocodev{}} for the best performing model \baseBcccoco{} on all 35 languages.
(\cocodev{} computed using machine-translated references).
}
\label{table:raw_cider_baseBcccoco}
\end{table}

\begin{table}[ht!]
\centering
\resizebox{0.6\linewidth}{!}{%
\rowcolors{3}{gray!15}{white}
\setlength{\tabcolsep}{12pt}
\begin{tabular}{cccS[table-format=1.3]S[table-format=1.3]} 
 \toprule
 \bf \bf {\multirow{2}{*}{\bf Lang.}}& {\bf CIDEr} & {\bf CIDEr} \\  
& {\bf \crossmodal} & {\bf \cocodev} \\
 \midrule
ar & 0.121 & 0.573 \\
bn & 0.139 & 0.623 \\
cs & 0.157 & 0.500 \\
da & 0.195 & 0.736 \\
de & 0.138 & 0.612 \\
el & 0.119 & 0.777 \\
en & 0.337 & 0.851 \\
es & 0.232 & 0.835 \\
fa & 0.180 & 0.816 \\
fi & 0.098 & 0.442 \\
fil & 0.215 & 0.951 \\
fr & 0.226 & 0.835 \\
he & 0.121 & 0.584 \\
hi & 0.112 & 0.718 \\
hr & 0.111 & 0.509 \\
hu & 0.107 & 0.491 \\
id & 0.187 & 1.000 \\
it & 0.184 & 0.783 \\
ja & 0.154 & 0.900 \\
ko & 0.169 & 0.787 \\
mi & 0.261 & 1.121 \\
nl & 0.235 & 0.697 \\
no & 0.242 & 0.768 \\
pl & 0.125 & 0.499 \\
pt & 0.222 & 0.852 \\
ro & 0.105 & 0.737 \\
ru & 0.111 & 0.588 \\
sv & 0.221 & 0.717 \\
sw & 0.191 & 0.702 \\
te & 0.112 & 0.497 \\
th & 0.253 & 0.856 \\
tr & 0.141 & 0.636 \\
uk & 0.091 & 0.631 \\
vi & 0.190 & 0.964 \\
zh & 0.110 & 0.695 \\
\bottomrule
\end{tabular}}
\caption{\textbf{CIDEr on \crossmodal{} and \cocodev{}} for the model \basegcoco{} on all 35 languages.
(\cocodev{} computed using machine-translated references).
}
\label{table:raw_cider_basegcoco}
\end{table}

\begin{table}[ht!]
\centering
\resizebox{0.6\linewidth}{!}{%
\rowcolors{3}{gray!15}{white}
\setlength{\tabcolsep}{12pt}
\begin{tabular}{cccS[table-format=1.3]S[table-format=1.3]} 
 \toprule
 \bf \bf {\multirow{2}{*}{\bf Lang.}}& {\bf CIDEr} & {\bf CIDEr} \\  
& {\bf \crossmodal} & {\bf \cocodev} \\
 \midrule
ar & 0.106 & 0.513 \\
bn & 0.133 & 0.555 \\
cs & 0.139 & 0.485 \\
da & 0.192 & 0.765 \\
de & 0.130 & 0.649 \\
el & 0.101 & 0.680 \\
en & 0.343 & 0.875 \\
es & 0.220 & 0.859 \\
fa & 0.155 & 0.766 \\
fi & 0.089 & 0.419 \\
fil & 0.185 & 0.858 \\
fr & 0.217 & 0.825 \\
he & 0.098 & 0.548 \\
hi & 0.111 & 0.624 \\
hr & 0.085 & 0.493 \\
hu & 0.096 & 0.451 \\
id & 0.167 & 0.943 \\
it & 0.168 & 0.770 \\
ja & 0.141 & 0.850 \\
ko & 0.152 & 0.716 \\
mi & 0.243 & 0.942 \\
nl & 0.232 & 0.704 \\
no & 0.230 & 0.736 \\
pl & 0.108 & 0.495 \\
pt & 0.202 & 0.843 \\
ro & 0.100 & 0.709 \\
ru & 0.089 & 0.581 \\
sv & 0.225 & 0.748 \\
sw & 0.151 & 0.640 \\
te & 0.099 & 0.426 \\
th & 0.226 & 0.802 \\
tr & 0.122 & 0.584 \\
uk & 0.081 & 0.560 \\
vi & 0.182 & 0.940 \\
zh & 0.099 & 0.656 \\
\bottomrule
\end{tabular}}
\caption{\textbf{CIDEr on \crossmodal{} and \cocodev{}} for the model \largegcoco{} on all 35 languages.
(\cocodev{} computed using machine-translated references).
}
\label{table:raw_cider_largegcoco}
\end{table}

\begin{table}[ht!]
\centering
\resizebox{0.6\linewidth}{!}{%
\rowcolors{3}{gray!15}{white}
\setlength{\tabcolsep}{12pt}
\begin{tabular}{cccS[table-format=1.3]S[table-format=1.3]} 
 \toprule
 \bf \bf {\multirow{2}{*}{\bf Lang.}}& {\bf CIDEr} & {\bf CIDEr} \\  
& {\bf \crossmodal} & {\bf \cocodev} \\
 \midrule
ar & 0.103 & 0.568 \\
bn & 0.107 & 0.594 \\
cs & 0.141 & 0.514 \\
da & 0.166 & 0.783 \\
de & 0.112 & 0.643 \\
el & 0.103 & 0.743 \\
en & 0.297 & 0.856 \\
es & 0.194 & 0.844 \\
fa & 0.152 & 0.793 \\
fi & 0.083 & 0.427 \\
fil & 0.181 & 0.878 \\
fr & 0.202 & 0.836 \\
he & 0.099 & 0.573 \\
hi & 0.098 & 0.671 \\
hr & 0.087 & 0.523 \\
hu & 0.087 & 0.478 \\
id & 0.149 & 0.947 \\
it & 0.153 & 0.780 \\
ja & 0.119 & 0.844 \\
ko & 0.134 & 0.760 \\
mi & 0.239 & 1.049 \\
nl & 0.219 & 0.705 \\
no & 0.197 & 0.743 \\
pl & 0.101 & 0.515 \\
pt & 0.176 & 0.832 \\
ro & 0.084 & 0.709 \\
ru & 0.077 & 0.586 \\
sv & 0.189 & 0.752 \\
sw & 0.158 & 0.681 \\
te & 0.097 & 0.463 \\
th & 0.210 & 0.830 \\
tr & 0.106 & 0.613 \\
uk & 0.081 & 0.580 \\
vi & 0.190 & 1.022 \\
zh & 0.087 & 0.659 \\
\bottomrule
\end{tabular}}
\caption{\textbf{CIDEr on \crossmodal{} and \cocodev{}} for the model \baseBcoco{} on all 35 languages.
(\cocodev{} computed using machine-translated references).
}
\label{table:raw_cider_basebcoco}
\end{table}

\end{document}